\documentclass[runningheads]{llncs}
\usepackage{graphicx}
\usepackage{comment}
\usepackage{amsmath,amssymb} 
\usepackage{color}
\usepackage[para]{footmisc}

\DeclareMathOperator*{\argmax}{\arg\!\max}

\def\ie{\emph{i.e.\ }}

\makeatletter
\newcommand{\printfnsymbol}[1]{%
	\textsuperscript{\@fnsymbol{#1}}%
}
\makeatother
\usepackage{hyperref}
\hypersetup{
    colorlinks=true,
    linkcolor=magenta,
    filecolor=magenta,      
    urlcolor=magenta,
}

\begin{document}
\pagestyle{headings}
\mainmatter
\def\ECCVSubNumber{672}  

\title{Multi-person 3D Pose Estimation in Crowded Scenes Based on Multi-View Geometry} 

\titlerunning{ECCV-20 submission ID \ECCVSubNumber} 
\authorrunning{ECCV-20 submission ID \ECCVSubNumber} 
\titlerunning{ECCV-20 submission ID \ECCVSubNumber} 
\institute{Paper ID \ECCVSubNumber}


\titlerunning{3D Crowd Pose Estimation Based on MVG}
%
\author{He Chen\thanks{Equal first author contribution.}\inst{1}, 
Pengfei Guo\printfnsymbol{1}\inst{1},
Pengfei Li\inst{1}, 
Gim Hee Lee\thanks{Jointly supervised this work.}\inst{2},
Gregory Chirikjian\printfnsymbol{2}\inst{2,1}}
\authorrunning{H Chen et al.}

%
\institute{The Johns Hopkins University, USA \and
National University of Singapore, Singapore\\
\email{\{hchen136, pguo4, pli32, gchirik1\}@jhu.edu}\\
\email{\{gimhee.lee@comp, mpegre@\}.nus.edu.sg}
}


\maketitle

\begin{abstract}
Epipolar constraints are at the core of feature matching and depth estimation in current multi-person multi-camera 3D human pose estimation methods. Despite the satisfactory performance of this formulation in sparser crowd scenes, its effectiveness is frequently challenged under denser crowd circumstances mainly due to two sources of ambiguity. The first is the mismatch of human joints resulting from the simple cues provided by the Euclidean distances between joints and epipolar lines. The second is
the lack of robustness from the naive formulation of the problem as a least squares minimization.
In this paper, we depart from the multi-person 3D pose estimation formulation, and instead reformulate it as crowd pose estimation. Our method consists of two key components: a graph model for fast cross-view matching, and a maximum a posteriori (MAP) estimator for the reconstruction of the 3D human poses. 
We demonstrate the effectiveness and superiority of our proposed method on four benchmark datasets. Our code is available at: \href{https://github.com/HeCraneChen/3D-Crowd-Pose-Estimation-Based-on-MVG}{https://github.com/HeCraneChen/3D-Crowd-Pose-Estimation-Based-on-MVG}.

\end{abstract}

\section{Introduction}
Fast 3D human pose estimation for crowded scenes is an important component in many computer vision applications such as autonomous driving, surveillance, and robotics~\cite{dinesh2018carfusion,fernando2018tracking,kubo2019programmable,li2018carn,li2020group,li20193d,stoll2011fast,vo2020self,xin2019theory}. However, recovering 3D human pose from crowded real-world setting is a challenging endeavor due to the inherent depth ambiguity caused by 2D to 3D backprojections, self-occlusions, and occlusions by other people in crowded scenes~\cite{baque2017deep,korman2018oatm,reddy2019occlusion}. 
A three-step process is commonly used in the multi-person multi-camera 3D pose estimation problem: 1) Detecting human body keypoints or parts in separate 2D views;  2) Matching people across different views; 3) Reconstructing 3D pose by triangulation. Unfortunately, the critical second step of matching people across different views is non-trivial. Well-known matching algorithms such as the Harris corner detector~\cite{harris1988combined} and the Scale Invariant Feature Transform (SIFT)~\cite{lowe1999object} give mostly wrong matches even after robust estimation with RANSAC~\cite{fischler1981random}. The problem is further aggravated in the third step when these unreliable matches are used in a vanilla triangulation algorithm to recover the 3D points.    

With the rapid development of deep learning, features are extracted more precisely and significant improvements are made for appearance-based feature matching across different viewpoints on the spatial level or different frames on the temporal level~\cite{liu2020extremely,sindagi2019multi,yew20183dfeat}. Despite the improvements, these methods are suboptimal for the task of people matching across multiple views in crowded scenarios. The reasons are threefold. Firstly, intra-class variation of human body appearance is relatively smaller than objects such as architectural features or graffiti paintings, and thus more outliers can result if the aforementioned methods are deployed directly. Secondly, dense feature matching across whole images is usually computationally inefficient for applications such as autonomous driving, where real-time is one of the primary concerns. Thirdly, appearance-based matching has a lower correctness criterion than people-based matching across multiple views. 
On the other hand, it is interesting to note that the level of occlusion in the same object can differ drastically among different views. Therefore, it is reasonable to trust the slightly occluded views more than the highly occluded views in the process of triangulation.

In this paper, we propose a 3D crowd human pose estimation method based on multi-view geometry. Specifically, we focus on overcoming the bottlenecks of multi-person 3D pose estimation and pushing it further to dense crowd 3D pose estimation. To this end, we propose the matching of feet across multiple views to improve the accuracy of body joint correspondences. We first modify a 2D pose estimation network, i.e. the joint-candidates single person pose estimation (SPPE)~\cite{li2019crowdpose} to include additional joints for the feet. Subsequently, we find the best matches of the feet across multiple views, and then extend the correspondences to the other joints using the kinematic chain of the human body. We cast the matching problem as a binary linear program and solve it efficiently with the Jonker-Volgenant algorithm~\cite{jonker1987shortest}. Finally, we improve the robustness of triangulation by formulating the problem as a maximum a posteriori (MAP) estimation that weighs the likelihood term with the uncertainty of the 2D joint observation and enforces a prior on the average bone lengths of the estimated 3D human poses. We evaluate our proposed method on four challenging benchmark datasets. Experimental results show that our method outperforms all existing algorithms on these datasets.

Our main contributions in this work are summarized as follows:
\begin{itemize}
\item Design a simple and efficient people matching mechanism based on feet assignment across different views, which is applicable for dense crowds.\vspace{1mm}
\item Propose a more robust triangulation for 3D crowd reconstruction using MAP estimation that accounts for the uncertainty of 2D joint detection and enforces the average 3D bone lengths. 
\vspace{1mm}
\item Define a problem of crowd 3D human pose estimation, and argue its existence as a separate problem from multi-person multi-camera 3D human pose estimation. 
\end{itemize}



\section{Related Work}
\paragraph {\bf{Single-Person Human Pose Estimation. }} A large amount of literature exists in this field due to the advancement of deep learning. We briefly summarize those for 3D human pose which are more closely related to this work. State-of-the-art methods can be divided into two categories, direct regression methods~\cite{chen20173d,jahangiri2017generating,mahendran20173d} and indirect regression methods based on heat maps~\cite{iskakov2019learnable,li2019generating,xzhou1}.  In~\cite{xzhou1},  a coarse-to-fine prediction scheme was developed by analyzing 3D human pose in a volumetric representation. Integral pose~\cite{sun2018integral} unifies the heat map representation and joint regression by replacing the non-differentiable \textit{argmax} with integral operation. Regardless of the good performance, learning 3D pose from a single image is still an  ill-posed problem. Instead of finding one exact solution,~\cite{li2019generating} developed a multimodal mixture density network, so that multiple feasible solutions are found before refining into one solution. The authors of~\cite{iskakov2019learnable} proposed a volumetric aggregation from intermediate 2D backbone feature maps and combines 3D information from multiple 2D views. The aforementioned methods obtained state of the art performance for single person 3D pose estimation, but unfortunately in the multi-person scenario, additional ambiguity makes these methods suboptimal.

\paragraph {\bf{Multi-Person Human Pose Estimation. }} Several recent works have focused on multi-person scenarios in problem formulation either based on monocular setting~\cite{wang2018robust} or multi-view setting~\cite{belagiannis20143d,Authors42,belagiannis2014multiple,bogo2016keep,Authors26,kadkhodamohammadi2018generalizable}. Results obtained from the multi-view setting are generally more precise due to the additional information. However, bottlenecks still exist in these multi-view based methods, \ie how to cope with the correspondence problem and how to make the triangulation of depth information sufficiently robust against noise. In~\cite{kadkhodamohammadi2018generalizable}, epipolar constraints are directly applied for people assignment among different views. This worked perfectly when people in the scene stand far away from each other. However, this constraint is likely to fail when the scenario gets crowded. For instance, if some epipolar line of a particular joint happens to pass through several other people, it is hard to make sure that no other joint is closer to the line than the correct matching joint. The authors of~\cite{Authors26} incorporated appearance cues by fusing re-identification with epipolar constraints. However, the two kinds of constraints are still independently considered. The 3D pictorial structure model~\cite{belagiannis20143d,Authors42} resolves ambiguities of mixed parts, occlusion, and false positives by building multi-view unary potentials, while at the same time integrating prior model by pairwise and ternary potential functions. This motivates our work in using MAP as a formulation to cope with measurement noise in triangulation process. 

Previous `multi-person' methods work on relatively sparse crowds. In~\cite{li2019crowdpose}, crowd pose estimation is firstly defined as a separate research field, but the problem is defined in 2D. When extending to 3D, more uncertainties are introduced. This encourages us to define the crowd pose estimation problem in 3D and explore a potential solution in this paper.

\paragraph {\bf{Feature Matching and Correspondence Problem. }} Feature correspondence in general raises stricter demand than feature matching due to the fact that both appearance and location need to be taken into consideration. In~\cite{campbell2017globally}, a globally-optimal inlier set cardinality maximization approach is proposed to jointly estimate optimal camera pose and optimal correspondences. \cite{windheuser2016convex} solves the correspondence problem between two images by defining energy function measuring data consistency and spatial regularity. In~\cite{duff2019plmp}, Point-Line Minimal Problems are thoroughly defined and analyzed. This provides a theoretical guidance to solve the specific problem of point line matching for the people assignment task. 
\section{Our Method}
\begin{figure}[!hbt]
\centering
\includegraphics[width=12cm]{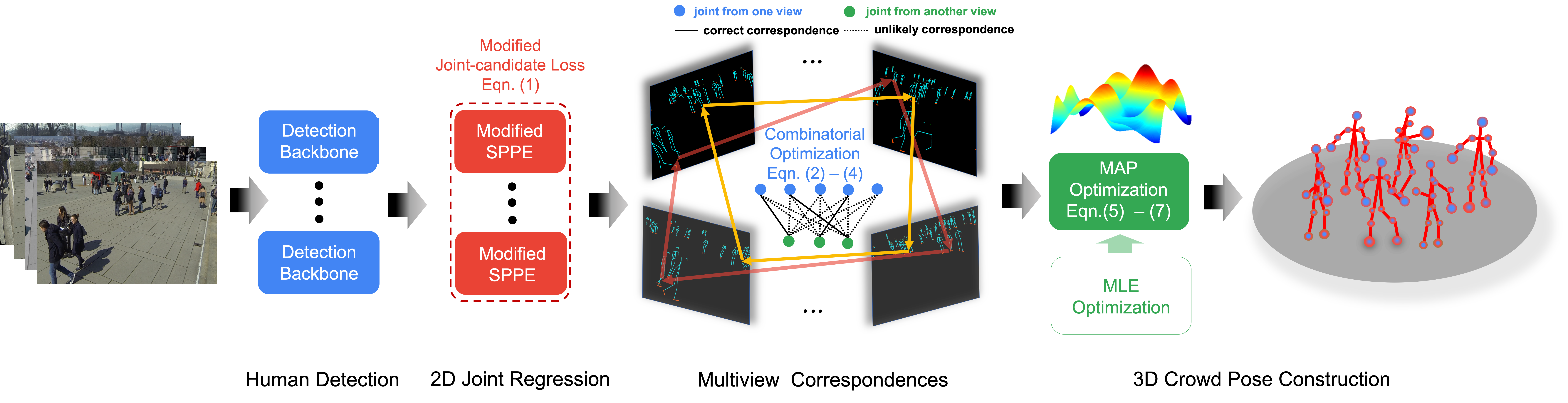}
\caption{The pipeline of our proposed approach. See text for more detail.
}
\label{fig:pipeline}
\end{figure}
\noindent Figure~\ref{fig:pipeline} shows an overview of our approach. Human bounding box proposals are first obtained by an off-the-shelf detection network, and then fed into a modified SPPE network (Sec.~\ref{sec:section_one}) to estimate the 2D joints. Subsequently, we get the multi-view joint correspondences by solving a combinatorial optimization problem via graph matching (Sec.~\ref{sec:section_two}). Finally, the 3D crowd poses are reconstructed using a MAP formulation
(Sec.~\ref{sec:section_three}) solved by the trust region method~\cite{conn2000trust}.


\subsection{2D Pose Estimation}
\label{sec:section_one}
We leverage on the recently proposed CrowdPose network~\cite{li2019crowdpose} trained on the CrowdPose Dataset~\cite{li2019crowdpose} for 2D pose detection on the input images. The CrowdPose network follows a top-down framework. It first detects the bounding boxes of individual persons using YOLOv3~\cite{redmon2018yolov3}, and then performs joint-candidate SPPE and a global maximum joints association algorithm to estimate the 2D joints. Similar to other 2D pose estimation methods, the accuracy of the joint detection drops as it moves farther away from the center of a person (i.e. the `hip' joint) despite the state-of-the-art performance of~\cite{li2019crowdpose} on the benchmark datasets.
As a result, detection of the `ankle' joints, which are usually used to represent feet, are especially noisy. To mitigate this problem, we follow~\cite{cao2018openpose} in adding 6 additional joints on the feet (3 on each foot) and modify the loss function of the network into the weighted sum of the mean square error (i.e. MSE[.,.]) from the body joints and the feet joints as follows:
\begin{equation}
    \mathcal{L} = 
    \frac{1}{I + 6\lambda}\left\{ \sum_{i=1}^{I} \text{MSE} \left[ \mathbf{P}_{h}^i,\mathbf{T}_{h}^i+\mu \mathbf{C}_{h}^i \right] +
    \lambda\sum_{i=I+1}^{I+7}\text{MSE}\left[\mathbf{P}_{h}^i,\mathbf{T}_{h}^i+\mu
    \mathbf{C}_{h}^i \right]\right\}.
\label{eq:5}
\end{equation}
$I$ stands for the number of joints of the body part excluding the 6 joints representing the feet (e.g. $I$ = 17 for MSCOCO~\cite{lin2014microsoft}). 
$\mathbf{P}_{h}^i$ and $\mathbf{T}_{h}^i$ represents the output heatmap and the heatmap of the target joints, respectively, for the $i^{\text{th}}$ joint of the $h^{\text{th}}$ person. $\mathbf{C}_{h}^i$ represents detections of the same joint type from other persons that might be within the bounding box of the $h^{\text{th}}$ person. 
We include $\mathbf{C}_{h}^i$ into the loss function to learn a multi-modal heatmap $\mathbf{P}_{h}^i$.
$\mu$ is the attention factor in the range of $[0,1]$ to control the extent of the contribution of $\mathbf{C}_{h}^i$, which we set to $0.5$ in all our implementations.  We set $\lambda > 1$, so that the 6 additional joints on the feet receive more attention during training. Our network is trained on the Human Foot Keypoint Dataset~\cite{cao2018openpose}.

\subsection{Multi-view Correspondence with Graph Matching}
\label{sec:section_two}
Previous methods~\cite{Authors26,kadkhodamohammadi2018generalizable} apply epipolar constraints to all joints in order to solve the correspondence problem. We argue that this can give a suboptimal solution when the crowd becomes denser. 
This is because the epipolar line that corresponds to a joint in one view is likely to pass through multiple joints in the other view for a crowded scene. Consequently, this ambiguity renders the Euclidean distance between the epipolar line and joints to be a less ideal metric.
We circumvent this challenge by casting the joint correspondence problem into a feet assignment problem. Specifically, we first establish the feet that belong to a same person across the multiple views, and then grow the joint correspondences from the feet using the kinematic chain of the human body.  

\begin{figure}[!hbt]
\centering
\includegraphics[width=12cm]{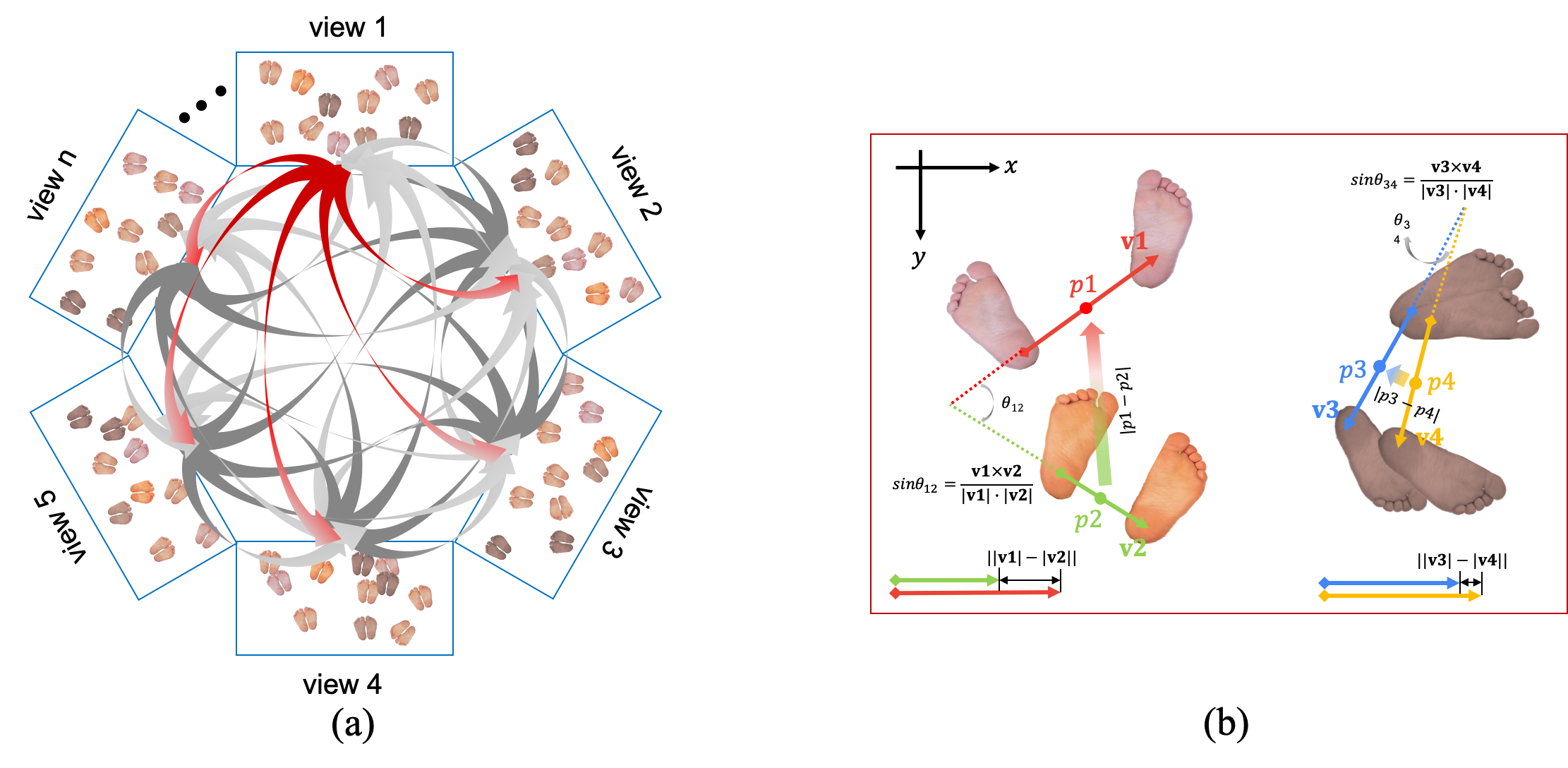}
\caption{People matching using feet assignment. (a) The matching process across $n$ views, and (b) visualization of edge cost defined in Eq.~\ref{eq:11}.}
\label{fig:2}
\end{figure}

\paragraph{\bf{Feet Assignment.}}
We propose to use feet assignment to realize people matching as shown in Figure~\ref{fig:2}(a). 
The core intuition is that prior information, appearance constraints, location constraints are naturally fused in such setting. We use the fact that at least one foot is on the ground when a person is walking
as the prior information. The detected joints of the feet as described in Sec.~\ref{sec:section_one} are used as the appearance information. To incorporate location constraints, we use the homographies between all view pairs to rectify the ground planes among different views into a common reference. We denote the homography between the ground planes of view $j$ and $k$ as $H_{j,k}$. Consequently, we can directly compare the joints of the feet across different views. We get $\geq 4$ point correspondences between the ground plane of each pair of view $j$ and $k$ to compute $H_{j,k}$.
It is interesting to note that applying the homography to all pixels in the image, we might get a twisted image which appear to be strange at first glance. This is based on the prior that this `the world is 3D'. However, if we change the prior into `the world is 2D', and treat everything as chalk art drawn on the ground, then everything in the rectified image starts to look reasonable. In this light, the problem of joint matching boils down to feet assignment.

\paragraph{\bf{Graph Building. }} 
A naive search for the optimal feet assignment is intractable due to the large combinatorial search space. To improve the efficiency of the search, we build a complete bipartite graph from the feet across two views and solve it as a linear assignment problem.
Let $\mathcal{V}_{j} = \{v_{ij} : \forall i \in \{1, \dotsc, a_{j}\} \}$
denote the set of pair-of-feet in view $j$. $v_{ij}$ is the detected pair-of-feet with index $i$ in view $j$, and $a_{j}$ is the total number of detected pair-of-feet in view $j$. 
We further denote the set of edges in the complete bipartite graph for the pair of views $j$ and $k$ as $\mathcal{E}_{j,k} = \{e_{l,m} : \forall l \in \{1, \dotsc, a_{j}\}, m \in \{1, \dotsc, a_{k}\}\}$. The complete bipartite graph for each pair of views can then be formally written as:
\begin{equation}
\mathcal{K}_{a_{j},a_{k}} = ((\mathcal{V}_{j}, \mathcal{V}_{k}),\mathcal{E}_{j,k}) ,\label{eq:6}
\end{equation}

\paragraph{\bf{Optimal Cross-view Matching.}} 
Based on this construction, our goal becomes finding a subgraph $\mathcal{G} \subset \mathcal{K}_{a_{j},a_{k}}$ by eliminating edges in the graph that represent the unlikely correspondences. We solve this edge elimination problem as a binary linear program that minimizes the total edge costs subjected to a set of linear constraints, i.e.  
\begin{equation} \label{eq:assignmentCost}
\begin{aligned}
\min_{\mathbf{d}} \quad & \sum_{l=1}^{a_j} \sum_{m=1}^{a_k} c_{l,m} \cdot d_{l,m}\\
\textrm{s.t.} \quad & \sum_{l=1}^{a_j} c_{l,m} \leq 1, \quad \sum_{m=1}^{a_k} c_{l,m} \leq 1,\\
  & \sum_{l=1}^{a_j} d_{l,m}=1, \quad \sum_{m=1}^{a_k} d_{l,m}=1, \quad \mathbf{d} \in \{0,1\}^{a_j \times a_k}. \\
\end{aligned}
\end{equation}
$d_{l,m} \in \mathbf{d}$ is a binary variable that represents the selection of the edge $e_{l,m}$ when it is equals to 1. $c_{l,m}$ is the cost of selecting the edge $e_{l,m}$, which we define as:
\begin{equation}
\begin{aligned}
c_{l,m} = k_1 \cdot \big| p_{l} - H_{j,k} \cdot p_{m} \big| 
 + k_2 \cdot \big| |\textbf{v}_{l}\rvert - \lvert\textbf{v}_{m}| \big|
 + k_3 \cdot \left(\frac{\textbf{v}_l \times \textbf{v}_{m}}{|\textbf{v}_{l}| \cdot |\textbf{v}_{m}|}\right) 
,\label{eq:11}
\end{aligned}
\end{equation}
where $p_{l}$ and $p_{m}$ respectively represents the location of two pairs of feet, $H_{j,k}$ represents the homography matrix between the two views $j$ and $k$, $\textbf{v}_{l}$ and $\textbf{v}_{m}$ represent vectors of strides. $k_1$, $k_2$, $k_3$ are hyper parameters to adjust the importance between the foot location, stride size, and stride direction. The metric is visualized in Figure~\ref{fig:2}(b).


\paragraph{\bf{Solver.}} We use the Jonker-Volgenant algorithm~\cite{jonker1987shortest} as the solver to find the solution to the two-view feet assignment problem formulated in Eq.~\ref{eq:assignmentCost}. We ensure consistency of the assignment across multiple views by resolving the conflict in the correspondences with priority given to edges with lower edge cost as defined in Eq.~\ref{eq:11}. 
A directed graph where the skeleton is a spanning union of disjoint cycles is obtained when the matching across $n$ views is successful. Our matching algorithm has a time complexity of $\mathcal{O}((2N)^3) = \mathcal{O}(8N^3)$, where $N$ is the number of persons per image. In contrast, the $O(n^4)$ implementation of Hungarian algorithm has a total time complexity of $\mathcal{O}((17N)^4)$ on 17 joints. Although the constant term is usually considered unimportant for time complexity analysis, it cannot be neglected in this study since $N < 30$ usually holds. Thus, our method is significantly faster.
\vspace{-1.0cm} 
\subsection{3D Crowd Pose Reconstruction}
\label{sec:section_three}
Under the assumption that the camera parameters are known, we can reconstruct the 3D human poses by triangulation of the joint correspondences across the multiple views obtained from the previous section.
One naive method of triangulation is to directly minimize the squared sum of perpendicular distances between the epipolar line and the detected joint. We refer to this naive method as the vanilla triangulation method. This is a classical method that works well in single person scenarios. However, in occluded scenes, the 2D joints are noisy and might have shifts of a few pixels. Consequently, this breaks the correspondence across multiple views and causes the 3D reconstructed points to be unreliable. 
We formulate a MAP optimization
to mitigate the problem from the unreliable correspondences, where we model the likelihood with the 2D measurement uncertainty and use the prior term to constrain the bone lengths of the estimated body poses.  

\paragraph{\bf{MAP Optimization.}}\label{sec:section_three} The ultimate goal of the proposed method is to estimate 3D coordinates of human joints. We formulate this as a MAP over the latent 3D poses $\mathbf{Q}$, i.e.
\begin{equation}
\mathbf{Q}_{\rm{MAP}} 
= \argmax_{\mathbf{Q}} \prod \limits_{i = 1}^N P(Q_i) \prod \limits_{j = 1}^M \prod \limits_{k = 1}^O P(q_{ijk} \mid \mathcal{P}_k, Q_{ij}),
\label{eq:1}
\end{equation}
where $N$ is the total number of persons in the scene, $M$ is the number of joints per person, and $O$ is the total number of camera views. $q_{ijk}$ is the $j^\textrm{th}$ 2D joint of the $i^\textrm{th}$ person in the $k^\textrm{th}$ camera view. $Q_{ij} \in Q_i$ is the $j^\textrm{th}$ 3D joint from the 3D pose $Q_i \in \mathbf{Q}$ of the $i^\textrm{th}$ person in the scene. $\mathcal{P}_k$ is the projection matrix of the $k^{\text{th}}$ camera. The likelihood term is given by the following Gaussian distribution:
\begin{equation}\label{eq:3}
    P(q_{ijk} \mid \mathcal{P}_k, Q_{ij}) = \frac {1}{2 \pi \sigma_{ijk}} \exp \left\{-\frac {{\left\|q_{ijk} - \alpha(\mathcal{P}_k, Q_{ij})\right\| ^2}}{2 {\sigma_{ijk}} ^2}\right\},
\end{equation}
where $\sigma_{ijk} = f(s_{bbox}^i,  s_{heatmap}^k, q_{ijk})$ is the uncertainty of the $j^\textrm{th}$ 2D joint $q_{ijk}$ computed from the bounding box $s_{bbox}^i$ of the $i^\textrm{th}$ person and the output heatmap of the image from the $k^\textrm{th}$ view. $\left\|q_{ijk} - \alpha(\mathcal{P}_k, Q_{ij})\right\|$ is the reprojection error computed from the 2D joint $q_{ijk}$ and the normalized coordinates of the 3D joint $Q_{ij}$ projected into the image of the $k^\textrm{th}$ view given by $\alpha(., .)$. The prior term is defined as: 
    \begin{equation}\label{eq:4}
        P(Q_{i}) = \prod \limits_{l=1}^{L} \frac {1}{2 \pi \sigma_l} \exp \left\{ -\frac{\left\| b^l_{\textrm{ref}} - b^l_i \right\|^2}{2\sigma_l^2} \right\}, 
    \end{equation}
   where $b_i^l$ represents the $l^\textrm{th}$ bone length between two 3D joints in the $i^{\textrm{th}}$ person, and $b_{\textrm{ref}}^l$ represents the average length of the $l^\textrm{th}$ bone. $L$ is the total number of bones in the human body representation. $\sigma_l$ is the standard deviation in the length of the $l^\textrm{th}$ bone. Intuitively, the prior term enforces the bone lengths of the estimated 3D human pose to be close to the average lengths.
   
\paragraph {\bf{Initialization and solver.}} We initialize the iterative MAP optimization with the vanilla triangulation. Subsequently, we use the trust region method~\cite{conn2000trust} as a solver for the MAP optimization.  In addition, we empirically observe that performing the maximum likelihood estimation (MLE) with the initialized values as an intermediate step before MAP improves the final estimation of the 3d human poses.

\section{Experiments}
We evaluate our proposed method on four public datasets. These datasets consist of scenarios that include autonomous driving and surveillance with challenging situations such as moving camera and heavy occlusions.

\subsection{Datasets}

\paragraph{\bf{LOEWENPLATZ}~\cite{Authors48}.} This is a dataset of driving recorder scenario captured in Zurich with two calibrated cameras. The dataset represents common scenarios that autonomous driving cars are likely to experience everyday. 
\paragraph{\bf{Chariot Mk I}~\cite{Authors43}.} This is a dataset captured by hand-held cameras. The cameras are moving and shaking, which resemble real-life scenarios from the perspective of the pedestrians.
\paragraph{\bf{Wildtrack}~\cite{Authors44}:} This dataset emulates surveillance scenarios with the set-up of 7 fixed cameras.
All cameras are fully calibrated, i.e. known intrinsics and extrinsics camera parameters. Occlusion is severe in each view of this dataset.
\paragraph{\bf{CMU Panoptic Dataset}~\cite{Authors47}:} This dataset is captured in a studio and provides precise 3D ground truth in MSCOCO~\cite{lin2014microsoft} format. In this paper, we evaluate the performance of our method quantitatively on the `Ultimatum' sequences with complete 3D human pose annotations. This sequence consists of relatively more active and complicated social scenarios for human pose estimation than other sequences. 
\begin{table*}[!ht]
\centering
\setlength{\tabcolsep}{1.0pt}
\scriptsize
\caption{Quantitative results for the Chariot Mk I, LOEWENPLATZ, and Wildtrack datasets using the evaluation metrics from MSCOCO~\cite{lin2014microsoft}.}
\begin{center}
\begin{tabular}{c|c|cccc|c|cccc} 
\hline
Chariot Mk I &\textbf{AP} & AP$^{50}$ & AP$^{75}$ & AP$^{M}$ &  AP$^{L}$ & \textbf{AR} & AR$^{50}$ & AR$^{75}$ & AR$^{M}$ &  AR$^{L}$\\
\hline\hline
Belagiannis \textit{et al.}~\cite{Authors42} & 48.1 & 64.8 & 59.3 & 63.7 & 64.6 & 58.1 & 62.7 & 55.9 & 54.4 &61.9\\
Dong \textit{et al.}~\cite{Authors26} & 69.3 & 87.4 & 73.6 & 77.5 & 75.4 & 71.9 & 87.5 & 81.7 & 78.1 &80.0\\
Ours w/  Vanilla Trigulation & 60.0 & 90.8 & 72.2 & 65.4 & 77.6 & 72.3 & 95.3 & 83.0 & 76.6 & 81.8 \\
Ours w/ Proposed MAP & \textbf{89.8} & \textbf{98.9} & \textbf{92.7} & \textbf{91.7} & \textbf{99.5} & \textbf{93.9} & \textbf{99.8} & \textbf{96.0} & \textbf{95.4} & \textbf{99.6} \\
\hline
LOEWENPLATZ &\textbf{AP} & AP$^{50}$ & AP$^{75}$ & AP$^{M}$ &  AP$^{L}$ & \textbf{AR} & AR$^{50}$ & AR$^{75}$ & AR$^{M}$ &  AR$^{L}$\\
\hline\hline
Belagiannis \textit{et al.}~\cite{Authors42}& 49.3 & 63.7 & 58.2 & 63.2 & 56.9 & 61.9 & 84.3 & 64.3 & 73.7 &55.3  \\
Dong \textit{et al.}~\cite{Authors26} & 62.1 & 88.3 & 63.5 & 61.3 & 72.5 & 80.3 & 87.2 & 77.9 & 81.7&84.6\\
Ours w/ Vanilla Trigulation & 66.7 & 93.8 & 73.1 & 71.6 & 84.4 & 78.2 & 96.7 & 84.5 & 80.1 &88.9 \\
Ours w/ Proposed Optimization & \textbf{81.8} & \textbf{97.1} & \textbf{88.7}& \textbf{83.3} & \textbf{90.8} & \textbf{88.9} & \textbf{98.5} & \textbf{93.5} & \textbf{90.0} & \textbf{94.4} \\
\hline
Wildtrack &\textbf{AP} & AP$^{50}$ & AP$^{75}$ & AP$^{M}$ &  AP$^{L}$ & \textbf{AR} & AR$^{50}$ & AR$^{75}$ & AR$^{M}$ &  AR$^{L}$\\
\hline\hline
Belagiannis \textit{et al.}~\cite{Authors42} & 44.1 & 53.4 & 46.0 & 19.4 & 47.8 & 64.1 & 79.1 & 61.4 & 20.9 &55.4 \\
Dong \textit{et al.}~\cite{Authors26} & 55.6 & 78.4 & 53.1 & 34.9 & 60.0& 73.4 & 87.8 & 68.1 & 38.1 & 77.6 \\
Ours w/ Vanilla Trigulation & 55.3 & 79.6 & 50.6 & 33.2 & 60.1 & 77.3 & 88.7 & 72.9 & 38.6 &78.4 \\
Ours w/ Proposed MAP & \textbf{70.0} & \textbf{90.2} & \textbf{73.6} & \textbf{44.7} & \textbf{76.4} & \textbf{78.3} & \textbf{93.6} & \textbf{82.4} & \textbf{55.5} & \textbf{83.7} \\
\hline
\end{tabular}
\end{center}
\label{tab:1}
\end{table*}
\vspace{-1.0cm} 
\subsection{Results}
\paragraph {\bf{Quantitative Results. }} We adopt the key point evaluation metrics of MSCOCO \cite{lin2014microsoft}, i.e. the average precision (AP), average recall (AR) and their variants. Specifically, the variants of AP and AR are specified by the Object Keypoint Similarity (OKS) that plays the same role as the Intersection over Union (IoU) in object detection. It measures the scale of the object, and the distance between predicted joints and ground truth points. 
The AP at OKS=.50:.05:.95 (primary challenge metric in MSCOCO~\cite{lin2014microsoft} competitions) is used to measure the reprojection errors. Table~\ref{tab:1} shows that our method outperforms the state-of-the-art algorithms on the Chariot Mk I, LOEWENPLATZ, and Wildtrack datasets using the evalution metrics from MSCOCO~\cite{lin2014microsoft}. Table~\ref{tab:2} shows the comparative performance for 2D key point detection of our modified body+foot candidate-joint SPPE network on the MSCOCO dataset~\cite{lin2014microsoft}. 
Our method achieves a comparable performance with the best performing~\cite{Authors14}. Furthermore, our method outperforms on AP@0.5:0.95 for medium objects, which is more valuable for our framework with the feet detection, matching and optimization stages.
CMU Panoptic Dataset provides the 3D ground truth. Therefore, we use two metrics, i.e. mean per joint position error (MPJPE) and percentage of correct parts (PCP) instead of the reprojection error for direct evaluation. The results are shown in Table~\ref{tab:CMU MPJPE} and Table~\ref{tab:CMU PCP}.
\begin{table}[!hbt]
\setlength{\tabcolsep}{4pt}
\scriptsize
\caption{Quantitative comparison of key point detection experiments on COCO body+foot validation set \cite{cao2018openpose}. 
}
\scriptsize
\begin{center}
\begin{tabular}{c|c|cccc} 
\hline
Method &\textbf{AP} & AP$^{50}$ & AP$^{75}$ & AP$^{M}$ &  A$P^{L}$\\
\hline\hline
GT Bbox + CPM~\cite{Authors45} & 62.7 & 86.0 & 69.3 & 58.5 & 70.6 \\
SSD~\cite{Authors07} + CPM~\cite{Authors45} & 52.7 & 71.1 & 57.2 & 47.0&64.2 \\
Cao \textit{et al.}~\cite{Authors14}& 65.3 &  \textbf{85.2} & 71.3 & 62.2 &  \textbf{70.7} \\
Ours & \textbf{65.3} & 80.1 & \textbf{72.2} & \textbf{74.1} &68.3  \\
\hline
\end{tabular}
\end{center}
\label{tab:2}
\end{table}
\begin{table}[!hbt]
\vskip-1cm
\setlength{\tabcolsep}{8pt}
\scriptsize
\caption{Quantitative results for the proposed method on different joints of human body in CMU Panoptic Dataset (Ultimatum sequences, four cameras) using MPJPE (mm).}
\scriptsize
\begin{center}
\begin{tabular}{c|c|cccccccccc} 
\hline
Metric &\textbf{Average} & Head & Shoulder & Elbow & Wrist &  Hip & Knee & Foot\\
\hline\hline
MPJPE & 50.0 & 45.1 & 43.6  & 55.6 & 60.7 & 25.3 & 53.2 & 66.0 \\
\hline
\end{tabular}
\end{center}
\label{tab:CMU MPJPE}
\end{table}
\begin{table}[!hbt]
\vskip-1cm
\setlength{\tabcolsep}{6pt}
\scriptsize
\caption{Quantitative results for the proposed method on different body parts in CMU Panoptic Dataset (Ultimatum sequences, four cameras) using the PCP metric.}
\scriptsize
\begin{center}
\begin{tabular}{c|c|ccccccc} 
\hline
Metric &\textbf{PCP}  & Head & Torso & Upper arms &  Lower arms & Upper legs & Lower legs\\
\hline\hline
percentage & 91.3 & 74.5 & 100.0 & 93.8 & 80.0 & 100.0 & 99.3 \\
\hline
\end{tabular}
\end{center}
\label{tab:CMU PCP}
\end{table}
\begin{figure*}[!hbt]
  \centering
  \includegraphics[width=12cm]{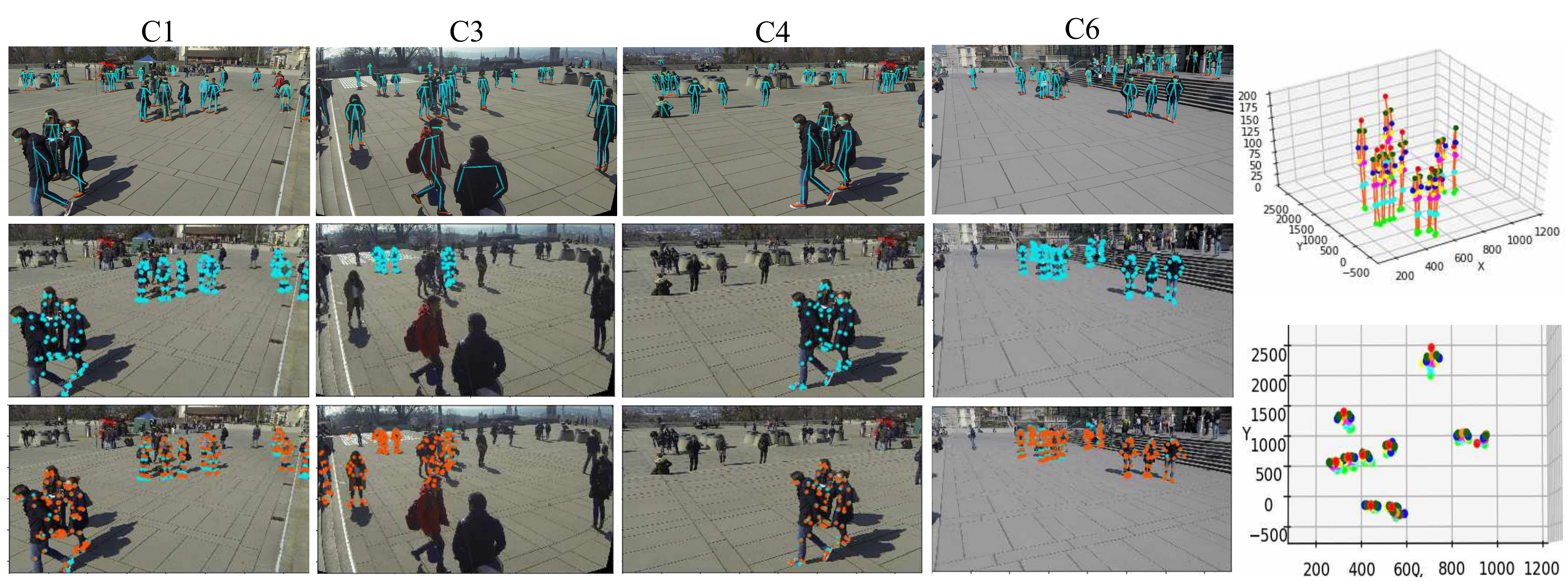}
  \caption{Qualitative results on Wildtrack dataset. (First four columns) First row shows results of our modified candidate joint SPPE with attention on the feet; Second row shows the ground truth 2D joints (blue dots); Third row shows the reprojection of our estimated 3D joints (orange dots) overlaid on the ground truths (blue dots).
  The last column shows the (top) estimated 3D crowd human poses and its (bottom) top view.
  }\label{fig:5}
\end{figure*}
\begin{figure*}[!hbt]
  \centering
  \includegraphics[width= 12cm]{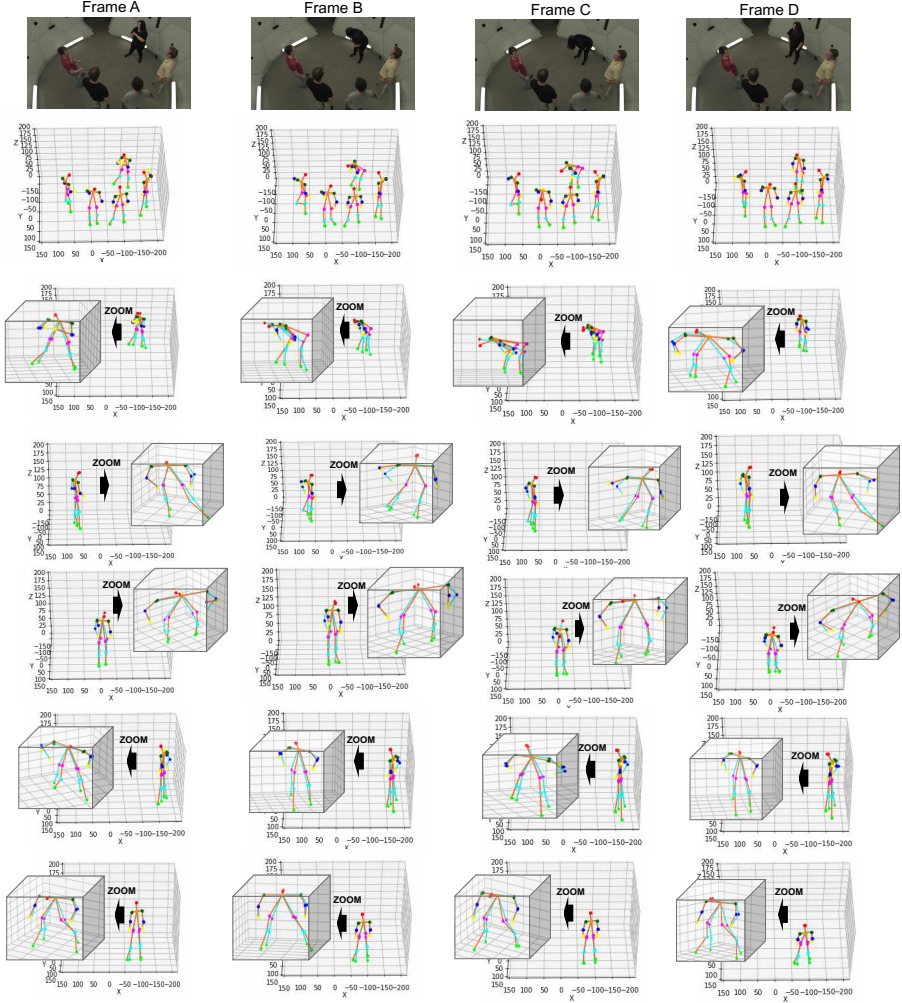}
  \caption{Qualitative results on 
  CMU Panoptic dataset. 
  The first row shows images from the 4 cameras in the setup.
  The second row shows 3D crowd pose. The third to seventh row visualize the estimated 3D pose of each person (orange skeleton) and its corresponding ground truth (blue skeleton).
  }
  \label{fig:6}
\end{figure*}
\begin{figure*}[!hbt]
  \centering
  \includegraphics[width= 12cm]{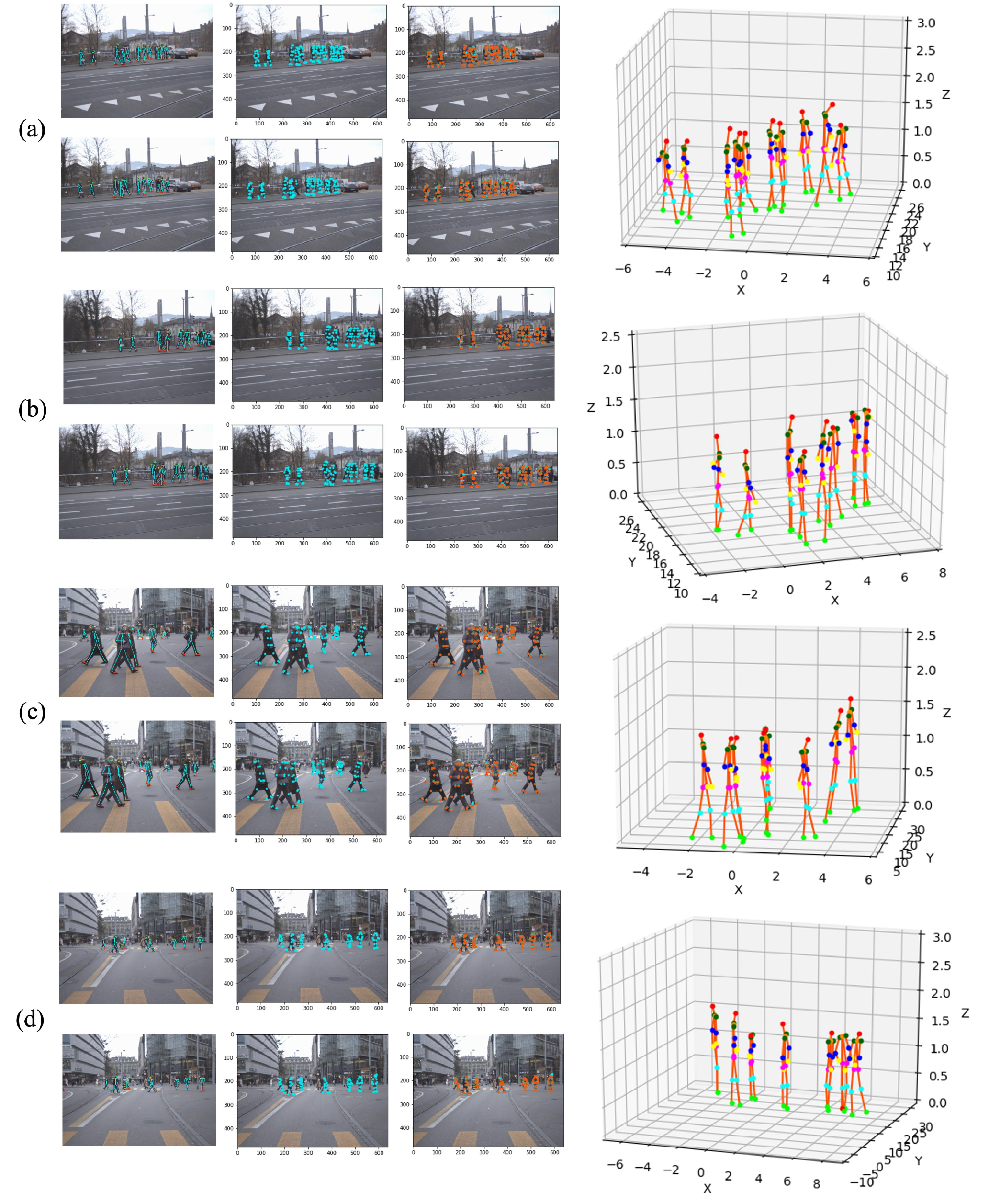}
  \caption{Qualitative results on the LOEWENPLATZ dataset. The right most column shows the estimated 3D poses of scene (a)-(d). 
  The first column shows the 2D skeletons detected by our modified SPPE network, the second column shows the ground truths of the 2D joints (blue dots), and the third column shows the reprojection of our estimated 3D joints (orange dots) and overlaid on the ground truths (blue dots).
  }
  \label{fig:8}
\end{figure*}
\begin{table}[!hbt]
\vskip-1cm
\caption{Ablation study of MLE as an intermediate step on WildTrack dataset.}
\setlength{\tabcolsep}{6pt}
\scriptsize
\begin{center}
\begin{tabular}{c|cccc} 
\hline
Method &ave & min&  max & var \\
\hline\hline
Ours w/o MLE & 64.75 & 18.06 & 316.7 & 50.69 \\
Ours w/ MLE & \textbf{38.55} & \textbf{2.18} & \textbf{219.29} & \textbf{27.52} \\
\hline
\end{tabular}
\end{center}
\label{tab:3}
\end{table}
\begin{table}[!hbt]
        \begin{minipage}{0.48\textwidth}
            \centering
            \scriptsize
            \caption{Foot keypoint analysis on COCO foot validation set.}
            \setlength{\tabcolsep}{1pt}
            \begin{tabular}{c|cc|cc}
            \hline
            Method             & \textbf{AP}   & \textbf{AR}  & AP$^{75}$ & AR$^{75}$ \\ \hline\hline
            Cao et al. {[}8{]} & 77.9 & 82.5 & 82.1 & 85.6                   \\ \hline
            Our                & 80.1 & 82.0 & 85.5 & 87.4                   \\ \hline
            \end{tabular}
            \label{tab:4}
        \end{minipage}
        \begin{minipage}{0.48\textwidth}
            \centering
            \scriptsize
            \caption{Evaluation of correspondence process on CMU Panoptic Dataset.}
            \setlength{\tabcolsep}{1pt}
            \begin{tabular}{c|ccccc} 
            \hline
            Dataset & RANSAC & EC & Ours \\
            \hline\hline
            Precision & 46.0 & 86.5 & 93.7 & \\
            \hline
            Time Complexity & $NA$ & $O((17N)^4)$ & $O((2N)^3)$& \\
            \hline
            \end{tabular}
            \label{tab:correspond}
        \end{minipage}
\end{table}

\begin{figure}[!hbt]
\centering
  \includegraphics[width= 9cm]{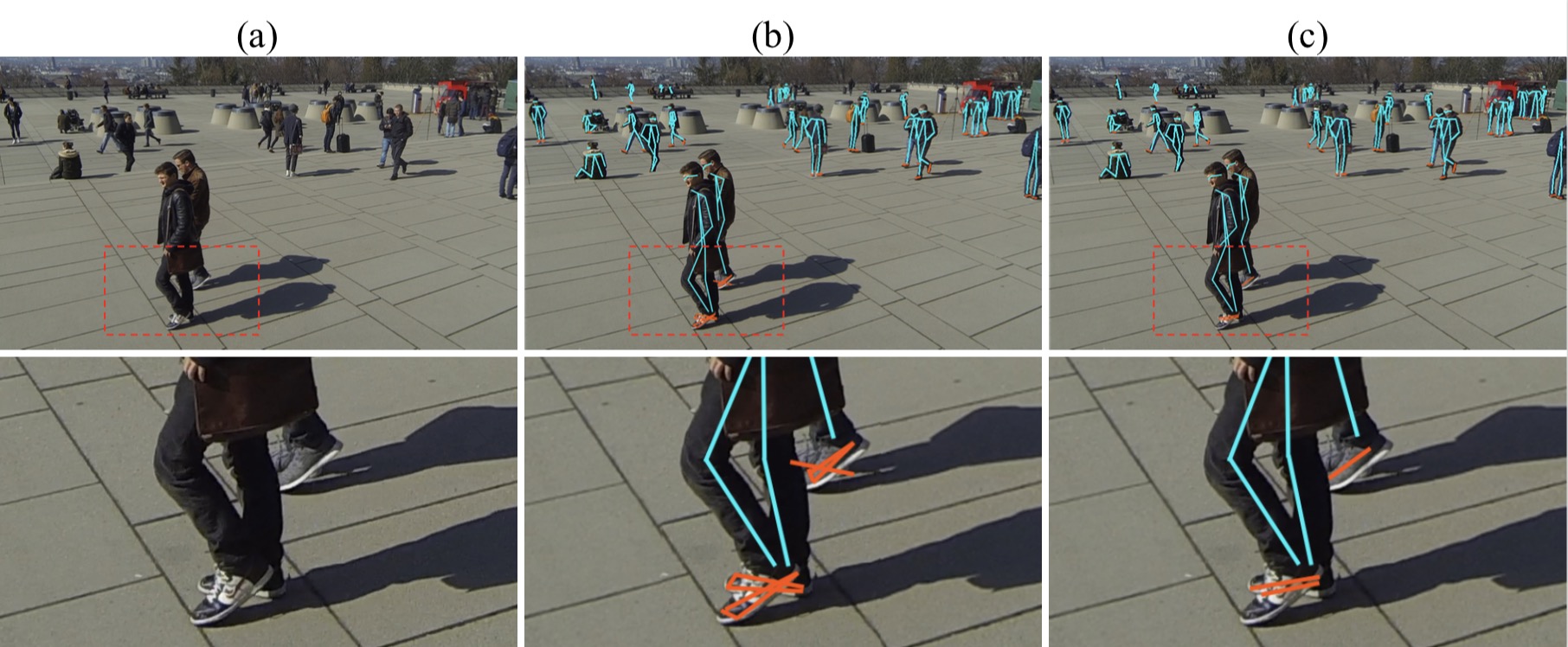}
  \caption{Qualitative demonstration of our proposed loss function in Eq.~\ref{eq:5}. The figure shows the (a) original image, and the pose estimation result (b) \textbf{with} and (c) \textbf{without} the loss term on the feet joints in Eq.~\ref{eq:5}. The second row shows the corresponding zoomed-in images.}  
  \label{fig:9}
\end{figure}
\begin{figure}[!hbt]
\centering
  \includegraphics[scale=0.3]{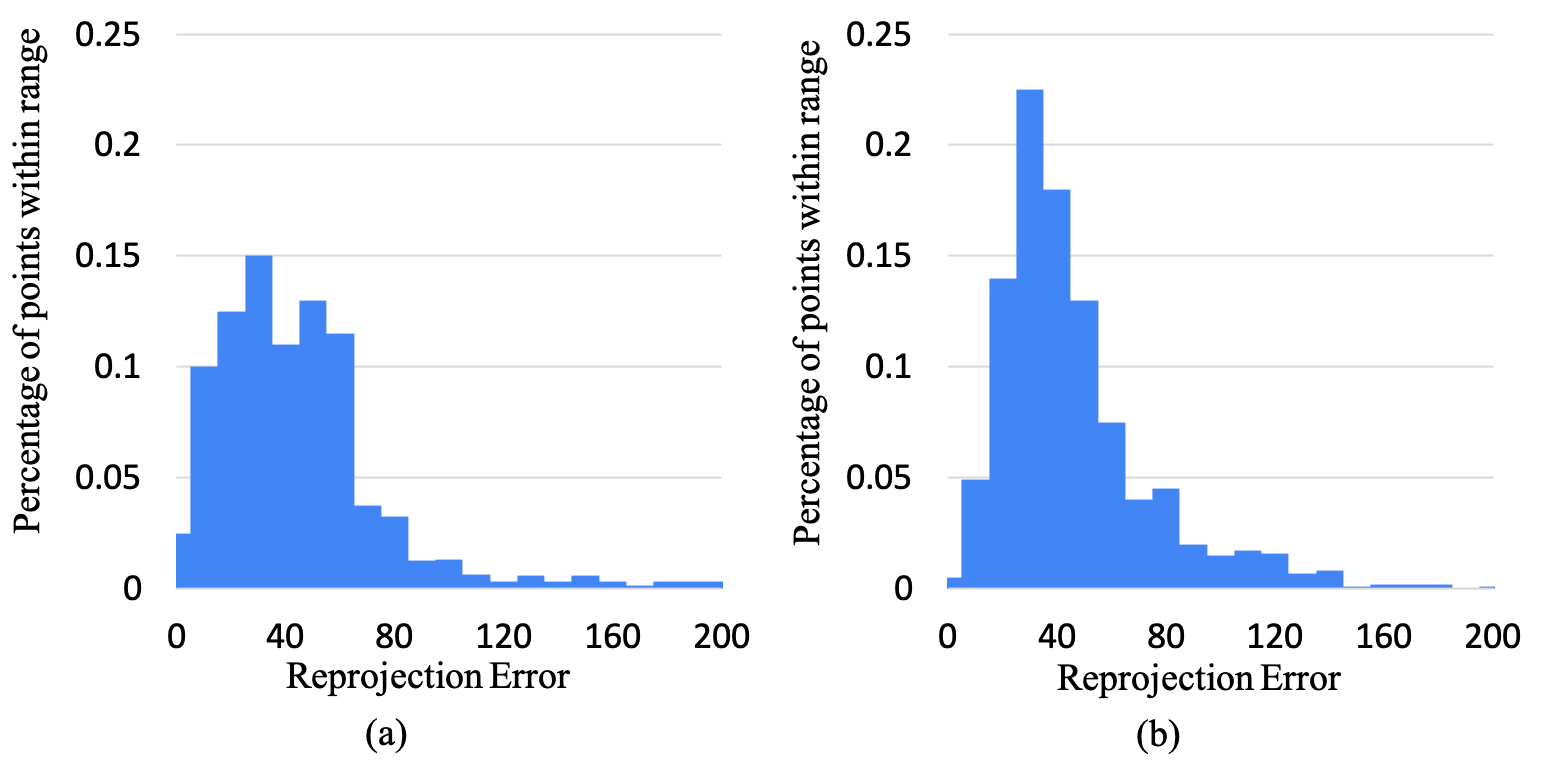}
  \caption{Error distributions:(a) without and (b) with MLE as an intermediate step.
  }
  \label{fig:7}
\end{figure}

\paragraph {\bf{Qualitative Results. }} Figure~\ref{fig:5},~\ref{fig:6}, and~\ref{fig:8} show the qualitative results on the Wildtrack~\cite{Authors44}, CMU Panoptic~\cite{Authors47}, and LOEWENPLATZ~\cite{Authors48} datasets, respectively. In Figure~\ref{fig:5}, our approach gives good quality 3D reconstructions of the human poses even when heavy occlusion happens in the crowded scene. To validate effectiveness of the proposed method, we choose crowded scenes with at least 5 people appearing in each frame
as shown Figure~\ref{fig:6}. We further show the qualitative visualizations of the estimated 3D human pose of several single persons from our method with the ground truth. Location information is used to match estimated pose with ground truth of each individual person. Orange represents estimated skeleton and blue represents ground truth. We zoom in each skeleton to clearly show details. As can be observed, the blue skeleton and orange skeleton has a slight offset. Nonetheless, this offset is in a tolerable range. In Figure~\ref{fig:8}, we evaluate our method under the setting of autonomous driving. The car went straight, turned left, and stopped at a crosswalk. We can see that our proposed method gives good 3D human pose estimations in different road scenes from a moving camera.

\paragraph {\bf{Ablation Study. }} We perform ablation studies to show the effectiveness of our proposed loss function Eq.~\ref{eq:5} for 2D pose estimation, and the MLE as an intermediate step. We define an error distance between the reprojection of a 3D point and its corresponding 2D ground truth for quantitative evaluation. 
Comparison is carried out between the results from MAP with and without MLE as an intermediate step on the WildTrack dataset.
In Table~\ref{tab:3}, we show the average, minimum, maximum, and variance of the reprojection error distance. Figure~\ref{fig:7} shows the histogram of error distribution in pixel unit. We can see that the smaller errors of the estimated 3D poses are obtained with the MLE as an intermediate step. Figure~\ref{fig:9} demonstrates the effectiveness of our proposed loss function Eq.~\ref{eq:5} for 2D pose estimation. As can be seen in the figure, our network detects the `big toe', `small toe' and `heel' instead of the usual `ankle' for the representation of a foot. The increased attention of the feet joints improves the estimation of the feet in highly occluded scene, and consequently facilities our matching algorithm. Comparison of the foot keypoints on the COCO foot validation set is shown in Table~\ref{tab:4}. To ablate the correspondence procedure, we conduct evaluations of correspondence process on the CMU Panoptic dataset in Table~\ref{tab:correspond}, where EC denotes Epipolar Constraint.
\section{Conclusions}
In this work, we propose a simple and effective approach for multi-person 3D pose estimation applicable to dense crowds. Matching of feet across multiple views improves the accuracy of body joint correspondences. A graph model is used for fast cross-view matching based on accurate estimation of foot joints. We cast the bipartite matching problem as a binary linear program and solve it efficiently with the Jonker-Volgenant algorithm. The robustness of triangulation is improved by using a MAP estimation that weighs the likelihood term with the uncertainty of the 2D joint observation and enforces a prior on the average bone lengths of the estimated 3D human poses. Experimental results show that our method outperforms all existing algorithms on four public datasets.

\paragraph{\bf{Acknowledgements. }} The authors would like to thank Yawei Li and Weixiao Liu for useful discussion. This work is supported in parts by the Office of Naval Research Award N00014-17-1-2142 and the Singapore MOE Tier 1 grant
R-252-000-A65-114.
\clearpage
%
%
\bibliographystyle{splncs04}
\bibliography{egbib}

\begin{thebibliography}{10}
\providecommand{\url}[1]{\texttt{#1}}
\providecommand{\urlprefix}{URL }
\providecommand{\doi}[1]{https://doi.org/#1}

\bibitem{baque2017deep}
Baqu{\'e}, P., Fleuret, F., Fua, P.: Deep occlusion reasoning for multi-camera
  multi-target detection. In: Proc. ICCV. pp. 271--279 (2017)

\bibitem{belagiannis20143d}
Belagiannis, V., Amin, S., Andriluka, M., Schiele, B., Navab, N., Ilic, S.:
  3{D} pictorial structures for multiple human pose estimation. In: Proc. CVPR.
  pp. 1669--1676 (2014)

\bibitem{Authors42}
Belagiannis, V., Amin, S., Andriluka, M., Schiele, B., Navab, N., Ilic, S.:
  {3D} pictorial structures revisited: Multiple human pose estimation. IEEE
  Trans. PAMI  \textbf{38}(10),  1929--1942 (2015)

\bibitem{belagiannis2014multiple}
Belagiannis, V., Wang, X., Schiele, B., Fua, P., Ilic, S., Navab, N.: Multiple
  human pose estimation with temporally consistent 3{D} pictorial structures.
  In: Proc. ECCV. pp. 742--754. Springer (2014)

\bibitem{bogo2016keep}
Bogo, F., Kanazawa, A., Lassner, C., Gehler, P., Romero, J., Black, M.J.: Keep
  it smpl: Automatic estimation of 3{D} human pose and shape from a single
  image. In: Proc. ECCV. pp. 561--578. Springer (2016)

\bibitem{campbell2017globally}
Campbell, D., Petersson, L., Kneip, L., Li, H.: Globally-optimal inlier set
  maximisation for simultaneous camera pose and feature correspondence. In:
  Proc. ICCV. pp. 1--10 (2017)

\bibitem{cao2018openpose}
Cao, Z., Hidalgo, G., Simon, T., Wei, S.E., Sheikh, Y.: Open{P}ose: realtime
  multi-person 2{D} pose estimation using {P}art {A}ffinity {F}ields. In: arXiv
  preprint arXiv:1812.08008 (2018)

\bibitem{Authors14}
Cao, Z., Simon, T., Wei, S.E., Sheikh, Y.: Realtime multi-person 2d pose
  estimation using part affinity fields. In: Proc. CVPR. pp. 7291--7299 (2017)

\bibitem{Authors44}
Chavdarova, T., Baqu{\'e}, P., Bouquet, S., Maksai, A., Jose, C., Bagautdinov,
  T., Lettry, L., Fua, P., Van~Gool, L., Fleuret, F.: Wildtrack: A multi-camera
  hd dataset for dense unscripted pedestrian detection. In: Proc. CVPR. pp.
  5030--5039 (2018)

\bibitem{chen20173d}
Chen, C.H., Ramanan, D.: 3{D} human pose estimation = 2{D} pose estimation+
  matching. In: Proc. CVPR. pp. 7035--7043 (2017)

\bibitem{conn2000trust}
Conn, A.R., Gould, N.I., Toint, P.L.: Trust region methods, vol.~1. Siam (2000)

\bibitem{dinesh2018carfusion}
Dinesh~Reddy, N., Vo, M., Narasimhan, S.G.: Carfusion: Combining point tracking
  and part detection for dynamic {3D} reconstruction of vehicles. In: Proc.
  CVPR. pp. 1906--1915 (2018)

\bibitem{Authors26}
Dong, J., Jiang, W., Huang, Q., Bao, H., Zhou, X.: Fast and robust multi-person
  {3D} pose estimation from multiple views. In: Proc. CVPR. pp. 7792--7801
  (2019)

\bibitem{duff2019plmp}
Duff, T., Kohn, K., Leykin, A., Pajdla, T.: Plmp-point-line minimal problems in
  complete multi-view visibility. In: Proc. ICCV. pp. 1675--1684 (2019)

\bibitem{Authors48}
Ess, A., Leibe, B., Schindler, K., Gool, L.V.: Robust multiperson tracking from
  a mobile platform. In: IEEE Trans. PAMI. pp. 1831--1846 (2009)

\bibitem{Authors43}
Ess, A., Leibe, B., Schindler, K., Van~Gool, L.: A mobile vision system for
  robust multi-person tracking. In: Proc. CVPR. pp.~1--8. IEEE (2008)

\bibitem{fernando2018tracking}
Fernando, T., Denman, S., Sridharan, S., Fookes, C.: Tracking by prediction: A
  deep generative model for mutli-person localisation and tracking. In: Proc.
  WACV. pp. 1122--1132. IEEE (2018)

\bibitem{fischler1981random}
Fischler, M.A., Bolles, R.C.: Random sample consensus: a paradigm for model
  fitting with applications to image analysis and automated cartography.
  Communications of the ACM  \textbf{24}(6),  381--395 (1981)

\bibitem{harris1988combined}
Harris, C.G., Stephens, M., et~al.: A combined corner and edge detector. In:
  Alvey vision conference. vol.~15, pp. 10--5244. Citeseer (1988)

\bibitem{iskakov2019learnable}
Iskakov, K., Burkov, E., Lempitsky, V., Malkov, Y.: Learnable triangulation of
  human pose. In: Proc. ICCV. pp. 7718--7727 (2019)

\bibitem{jahangiri2017generating}
Jahangiri, E., Yuille, A.L.: Generating multiple diverse hypotheses for human
  {3D} pose consistent with 2d joint detections. In: Proc. ICCVW. pp. 805--814
  (2017)

\bibitem{jonker1987shortest}
Jonker, R., Volgenant, A.: A shortest augmenting path algorithm for dense and
  sparse linear assignment problems. Computing  \textbf{38}(4),  325--340
  (1987)

\bibitem{Authors47}
Joo, H., Liu, H., Tan, L., Gui, L., Nabbe, B., Matthews, I., Kanade, T.,
  Nobuhara, S., Sheikh, Y.: Panoptic studio: A massively multiview system for
  social motion capture. In: Proc. ICCV (2015)

\bibitem{kadkhodamohammadi2018generalizable}
Kadkhodamohammadi, A., Padoy, N.: A generalizable approach for multi-view {3D}
  human pose regression. arXiv preprint arXiv:1804.10462  (2018)

\bibitem{korman2018oatm}
Korman, S., Milam, M., Soatto, S.: Oatm: Occlusion aware template matching by
  consensus set maximization. In: Proc. CVPR. pp. 2675--2683 (2018)

\bibitem{kubo2019programmable}
Kubo, H., Jayasuriya, S., Iwaguchi, T., Funatomi, T., Mukaigawa, Y.,
  Narasimhan, S.G.: Programmable non-epipolar indirect light transport: Capture
  and analysis. IEEE Trans. VCG  (2019)

\bibitem{li2019generating}
Li, C., Lee, G.H.: Generating multiple hypotheses for {3D} human pose
  estimation with mixture density network. In: Proc. CVPR. pp. 9887--9895
  (2019)

\bibitem{li2019crowdpose}
Li, J., Wang, C., Zhu, H., Mao, Y., Fang, H.S., Lu, C.: Crowdpose: Efficient
  crowded scenes pose estimation and a new benchmark. In: Proc. CVPR. pp.
  10863--10872 (2019)

\bibitem{li2018carn}
Li, Y., Agustsson, E., Gu, S., Timofte, R., Van~Gool, L.: {CARN}: Convolutional
  anchored regression network for fast and accurate single image
  super-resolution. In: Proc. ECCV. pp.~0--0 (2018)

\bibitem{li2020group}
Li, Y., Gu, S., Mayer, C., Van~Gool, L., Timofte, R.: Group sparsity: The hinge
  between filter pruning and decomposition for network compression. In: Proc.
  CVPR (2020)

\bibitem{li20193d}
Li, Y., Tsiminaki, V., Timofte, R., Pollefeys, M., Van~Gool, L.: 3{D}
  appearance super-resolution with deep learning. In: Proc. CVPR. pp.
  9671--9680 (2019)

\bibitem{lin2014microsoft}
Lin, T.Y., Maire, M., Belongie, S., Hays, J., Perona, P., Ramanan, D.,
  Doll{\'a}r, P., Zitnick, C.L.: Microsoft coco: Common objects in context. In:
  Proc. ECCV. pp. 740--755. Springer (2014)

\bibitem{Authors07}
Liu, W., Anguelov, D., Erhan, D., Szegedy, C., Reed, S., Fu, C.Y., Berg, A.C.:
  Ssd: Single shot multibox detector. In: Proc. ECCV. pp. 21--37. Springer
  (2016)

\bibitem{liu2020extremely}
Liu, X., Zheng, Y., Killeen, B., Ishii, M., Hager, G.D., Taylor, R.H.,
  Unberath, M.: Extremely dense point correspondences using a learned feature
  descriptor. In: Proc. CVPR (2020)

\bibitem{lowe1999object}
Lowe, D.G.: Object recognition from local scale-invariant features. In: Proc.
  ICCV. vol.~2, pp. 1150--1157 (1999)

\bibitem{mahendran20173d}
Mahendran, S., Ali, H., Vidal, R.: {3D} pose regression using convolutional
  neural networks. In: Proc. ICCVW. pp. 2174--2182 (2017)

\bibitem{xzhou1}
Pavlakos, G., Zhou, X., Derpanis, K.G., Daniilidis, K.: Coarse-to-fine
  volumetric prediction for single-image {3D} human pose. In: Proc. CVPR (2017)

\bibitem{reddy2019occlusion}
Reddy, N.D., Vo, M., Narasimhan, S.G.: Occlusion-{N}et: {2D/3D} occluded
  keypoint localization using graph networks. In: Proc. CVPR. pp. 7326--7335
  (2019)

\bibitem{redmon2018yolov3}
Redmon, J., Farhadi, A.: Yolov3: An incremental improvement. arXiv preprint
  arXiv:1804.02767  (2018)

\bibitem{sindagi2019multi}
Sindagi, V.A., Patel, V.M.: Multi-level bottom-top and top-bottom feature
  fusion for crowd counting. In: Proc. ICCV. pp. 1002--1012 (2019)

\bibitem{stoll2011fast}
Stoll, C., Hasler, N., Gall, J., Seidel, H.P., Theobalt, C.: Fast articulated
  motion tracking using a sums of gaussians body model. In: Proc. ICCV. pp.
  951--958. IEEE (2011)

\bibitem{sun2018integral}
Sun, X., Xiao, B., Wei, F., Liang, S., Wei, Y.: Integral human pose regression.
  In: Proc. ECCV. pp. 529--545 (2018)

\bibitem{vo2020self}
Vo, M., Yumer, E., Sunkavalli, K., Hadap, S., Sheikh, Y., Narasimhan, S.G.:
  Self-supervised multi-view person association and its applications. IEEE
  Trans. PAMI  (2020)

\bibitem{wang2018robust}
Wang, C., Wang, Y., Lin, Z., Yuille, A.L.: Robust {3D} human pose estimation
  from single images or video sequences. IEEE Trans. PAMI  \textbf{41}(5),
  1227--1241 (2018)

\bibitem{Authors45}
Wei, S.E., Ramakrishna, V., Kanade, T., Sheikh, Y.: Convolutional pose
  machines. In: Proc. CVPR. pp. 4724--4732 (2016)

\bibitem{windheuser2016convex}
Windheuser, T., Cremers, D.: A convex solution to spatially-regularized
  correspondence problems. In: Proc. ECCV. pp. 853--868. Springer (2016)

\bibitem{xin2019theory}
Xin, S., Nousias, S., Kutulakos, K.N., Sankaranarayanan, A.C., Narasimhan,
  S.G., Gkioulekas, I.: A theory of fermat paths for non-line-of-sight shape
  reconstruction. In: Proc. CVPR. pp. 6800--6809 (2019)

\bibitem{yew20183dfeat}
Yew, Z.J., Lee, G.H.: {3DF}eat-{N}et: Weakly supervised local {3D} features for
  point cloud registration. In: Proc. ECCV. pp. 630--646. Springer (2018)

\end{thebibliography}
\end{document}